\documentclass{article}

\usepackage{microtype}
\usepackage{graphicx}
\usepackage{subfigure}
\usepackage{booktabs} % for professional tables

\usepackage{hyperref}

\usepackage[accepted]{icml2024}

\usepackage{amsmath}
\usepackage{amssymb}
\usepackage{mathtools}
\usepackage{amsthm}

\usepackage[capitalize,noabbrev]{cleveref}

\usepackage{amsfonts,amsthm}
\usepackage{times}
\usepackage{latexsym}
\usepackage{comment}
\usepackage{graphicx}
\usepackage{xcolor}
\usepackage{color,soul}
\usepackage{booktabs}
\usepackage{mathtools}
\usepackage{linguex}
\usepackage{subfigure}
\usepackage{array}
\usepackage{float}
\usepackage{bm}
\usepackage{arydshln}
\usepackage{tikz}
\usepackage{multirow}
\usepackage{pmboxdraw}
\usepackage{breakurl}
\theoremstyle{plain}

\theoremstyle{definition}

\theoremstyle{remark}

\usepackage[textsize=tiny]{todonotes}

\usepackage{comment}
\usepackage{amsmath,amsfonts,bm}

\def\eqref#1{equation~\ref{#1}}
\def\1{\bm{1}}

\def\vb{{\bm{b}}}

\def\vu{{\bm{u}}}
\def\vv{{\bm{v}}}

\def\vx{{\bm{x}}}
\def\vy{{\bm{y}}}
\def\vz{{\bm{z}}}

\DeclareMathAlphabet{\mathsfit}{\encodingdefault}{\sfdefault}{m}{sl}
\SetMathAlphabet{\mathsfit}{bold}{\encodingdefault}{\sfdefault}{bx}{n}

\newcommand\numberthis{\addtocounter{equation}{1}\tag{\theequation}}

\icmltitlerunning{Information Flow Routes: Automatically Interpreting Language Models at Scale}

\let\svthefootnote\thefootnote
\begin{document}

\twocolumn[
\icmltitle{Information Flow Routes:\\ Automatically Interpreting Language Models at Scale}




\begin{icmlauthorlist}
\icmlauthor{Javier Ferrando$^{\ast}$}{upc}
\icmlauthor{Elena Voita}{meta}
\end{icmlauthorlist}

\icmlaffiliation{upc}{Universitat Politècnica de Catalunya}
\icmlaffiliation{meta}{FAIR at Meta AI}

\icmlcorrespondingauthor{Javier Ferrando}{javier.ferrando.monsonis@upc.edu}

\icmlkeywords{Machine Learning, ICML}

\vskip 0.3in
]



\printAffiliationsAndNotice{}  % leave blank if no need to mention equal contribution

\begin{abstract}
Information flows by routes inside the network via mechanisms implemented in the model. These routes can be represented as graphs where nodes correspond to token representations and edges to computations. We automatically build these graphs in a top-down manner, for each prediction leaving only the most important nodes and edges. In contrast to the existing workflows relying on activation patching, we do this through attribution: this allows us to efficiently uncover existing circuits with just a single forward pass. Unlike with patching, we do not need a human to carefully design prediction templates, and we can extract information flow routes for any prediction (not just the ones among the allowed templates). As a result, we can analyze model behavior in general, for specific types of predictions, or different domains. We experiment with Llama~2 and show that the role of some attention heads is overall important, e.g. previous token heads and subword merging heads. Next, we find similarities in Llama~2 behavior when handling tokens of the same part of speech. Finally, we show that some model components can be specialized on domains such as coding or multilingual texts.\footnote{The proposed method is implemented in: \url{https://github.com/facebookresearch/llm-transparency-tool}.}
\end{abstract}

\let\thefootnote\relax\footnote{$^{\ast}$Work done during an internship at Meta AI.}
\addtocounter{footnote}{-1}\let\thefootnote\svthefootnote
\vspace{-25pt}
\section{Introduction}
Current state-of-the-art LMs are built on top of the Transformer architecture~\citep{NIPS2017_3f5ee243,NEURIPS2020_1457c0d6,opt_lm,chowdhery2022palm,touvron2023llama,touvron2023llama2}. Inside the model, each representation evolves from the current input token embedding to the final representation used to predict the next token. This evolution happens through additive updates coming from attention and feed-forward blocks. The resulting stack of same-token representations is usually referred to as ``residual stream''~\citep{elhage2021mathematical}, and the overall computation inside the model can be viewed as a sequence of residual streams connected through layer blocks. Formally, we can see it as a graph where nodes correspond to token representations and edges to operations inside the model (attention heads, feed-forward layers, etc.).

\begin{figure}[!t]
	\begin{centering}
	\includegraphics[width=0.36\textwidth]{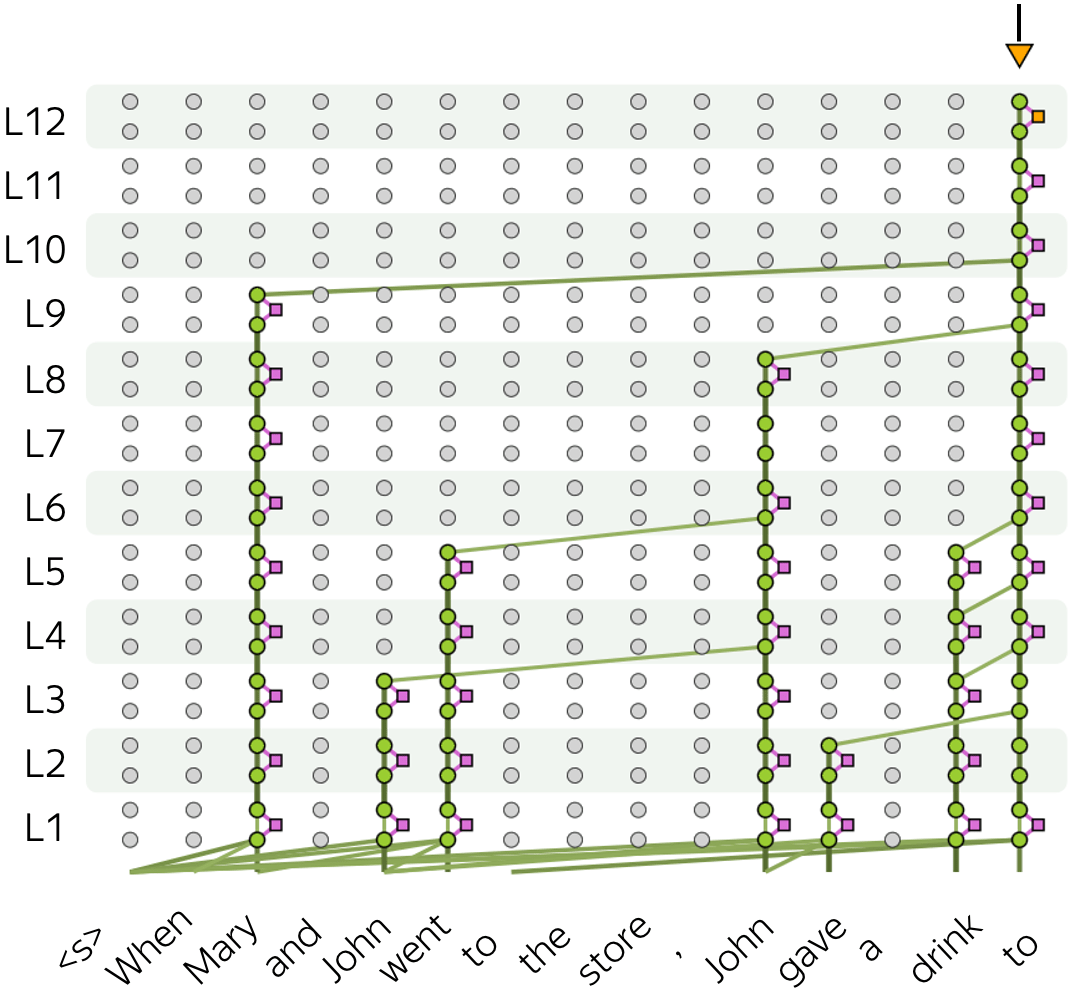}
	\caption{The important information flow routes for a token (Mary) prediction. GPT2-Small, $\tau = 0.04$.}
	\label{fig:ioi_gpt2_small_0.04}
	\end{centering}
 \vspace{-2ex}
\end{figure}

While during a forward pass all the edges are present, 
computations important for each prediction are likely to form a small portion of the original graph~(\citealp{voita-etal-2019-analyzing,wang2023interpretability,hanna2023does}, among others).
We extract this important subgraph in a top-down manner by tracing information back through the network and, at each step, leaving only edges that were relevant~(Figure~\ref{fig:ioi_gpt2_small_0.04}). To understand which edges are important, we rely on an attribution method~\cite{ferrando-etal-2022-measuring} and refuse from activation patching\footnote{Originally introduced by~\citet{causal_mediation_bias} to analyze LMs through the lens of causal mediation analysis.}, typical for the existing mechanistic interpretability workflows~\cite{wang2023interpretability,hanna2023does,conmy2023automated, stolfo2023understanding}. Firstly, patching requires human efforts to create templates and contrastive examples, thus is only applicable to a few pre-defined templates. Secondly, explaining a single prediction demands a substantial number of interventions (patches). Given that each intervention needs a forward pass, studying large models becomes increasingly impractical. In contrast, our method is about 100 times faster.

In the experiments, we first show that our information flow routes rely on the same task-specific attention heads found in patching circuits~\cite{wang2023interpretability,hanna2023does}. However, our method is more versatile and informative than patching: it can evaluate the importance of model components both (i)~overall for a prediction, and (ii)~compared to a contrastive example (i.e., patching setting). What is more, we argue that patching is fragile: its results can vary depending on the choice of the contrastive template (i.e., human preference).

\begin{figure}[!t]
	\begin{centering}
	\includegraphics[width=0.48\textwidth]{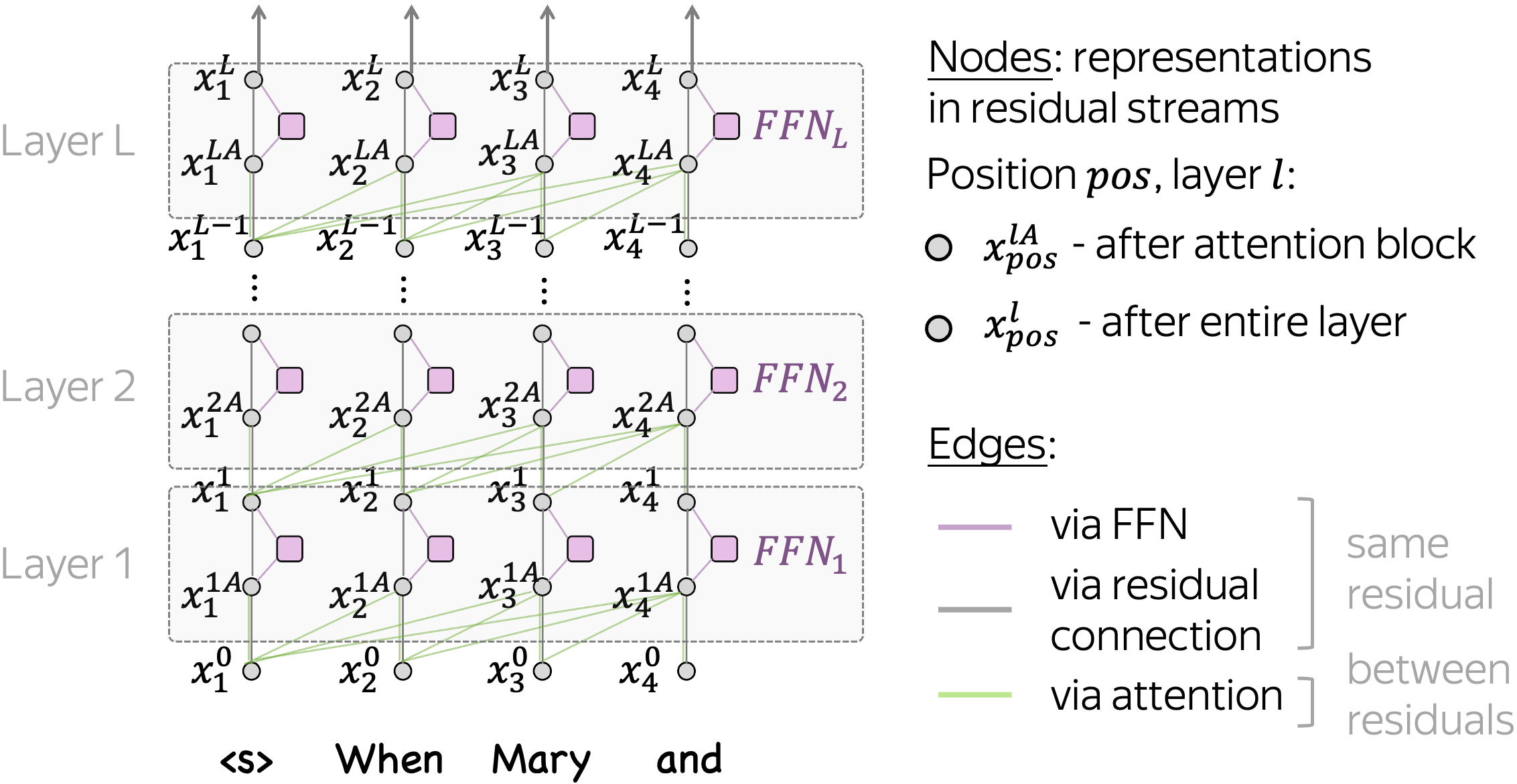}
	\caption{Full information flow graph.}
	\label{fig:graph_main}
	\end{centering}
 \vspace{-2ex}
\end{figure}

Next, we come to the settings unreachable to patching, i.e. the broad set of predictions and general importance. We experiment with Llama 2 and show that some attention head functions are overall important, e.g. previous token heads and subword merging heads. Next, we find that information inside Llama 2 flows similarly when handling tokens of the same part of speech. Finally, we show that some model components are specialized on domains such as coding or multilingual texts: they are active for these domains and not active otherwise.

Overall, our contributions are as follows:
\begin{itemize}
    \item we propose to explain predictions of transformer LMs via information flow routes;
    \item compared to patching circuits, our method is (i)~applicable to any prediction, (ii)~more informative, and (iii)~around 100 times faster;
    \item we analyze the information flow of Llama 2 and find model components that are (i)~generally important, and (ii)~specific to domains.
\end{itemize}

\section{Extracting Information Flow Routes}

\Cref{fig:graph_main} illustrates computations inside a Transformer LM along with the intermediate representations after each block.

\begin{figure}[!t]
	\begin{centering}
\includegraphics[width=0.45\textwidth]{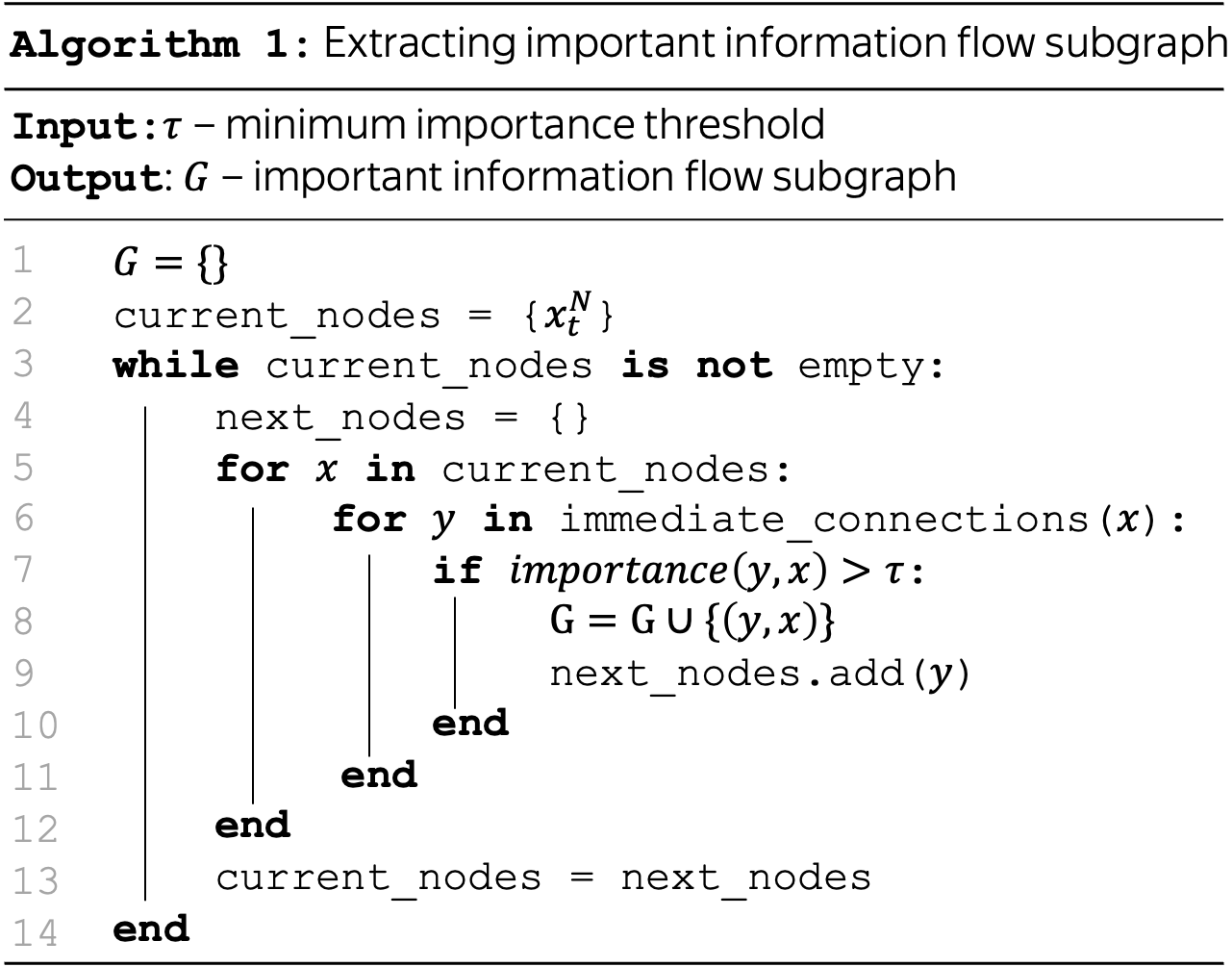}
	\caption{General-case algorithm for extracting the important subgraph, the information flow routes, from the full information graph~(\Cref{fig:graph_main}).}
	\label{fig:algorithm}
	\end{centering}
\end{figure}

\paragraph{Information flow graph.}
In the graph (i)~nodes correspond to token representations and (ii)~edges to operations inside the network moving information across nodes. Specifically, $x_{pos}^{lA}$ and $x_{pos}^{l}$ are representations of the token at position $pos$ after the attention block in layer $l$ or the entire layer, respectively. Each $x_{1}^{l-1}$, ..., $x_{pos}^{l-1}$ is connected to $x_{pos}^{lA}$ via attention edges (and via a residual stream edge from $x_{pos}^{l-1}$ to $x_{pos}^{lA}$), $x_{pos}^{lA}$ is connected to $x_{pos}^{l}$ via two edges: the FFN output and the residual stream.

\paragraph{Extracting the important subgraph.}
While during a forward pass all edges in \Cref{fig:graph_main} are present, computations relevant for each prediction are likely to form a small portion of the original graph~(\citealp{voita-etal-2019-analyzing,wang2023interpretability,hanna2023does}, etc).
We extract this important subgraph, the \textit{information flow routes}, in a top-down manner by tracing information back through the network~(\Cref{fig:algorithm}).
We start from a single node~-- the representation on top of the residual stream. Then, we go over immediately connected lower nodes and, if important at this step, we add them to the subgraph along with the corresponding edges. The algorithm requires setting a threshold $\tau$ for the minimum edge importance.

\paragraph{Importance via attribution, not patching.} To complete the proposed algorithm~(\Cref{fig:algorithm}), we need to specify how to compute the importance of an edge. While lately it has become typical to use patching~(see \Cref{apx:background_patching} for a more formal description)~\cite{wang2023interpretability,hanna2023does,conmy2023automated,docstring}, \textit{instead of patching, we choose to use attribution}. This choice is crucial for our work and makes our method about 100 times faster than alternatives while (i)~being able to recover previously discovered circuits, (ii)~doing this in a more versatile manner, and (iii)~leading to new observations. Next, we explain the specific attribution method we use.

\subsection{Evaluating Edge Importance}
\label{sect:edge_importance_full}

For the attribution method, we adopt ideas from ALTI~(Aggregation of Layer-Wise Token-to-Token Interactions)~\citep{ferrando-etal-2022-measuring}. While ALTI propagates attributions throughout the entire model, we only use its definition of contributions between connected nodes. We choose this method due to its simplicity, ease of implementation, and demonstrated effectiveness in practical applications, e.g. detecting hallucinations in neural machine translation~\cite{dale-etal-2023-detecting,dale-etal-2023-HalOmi,guerreiro2023hallucinations}.

\subsubsection{Importance in General Case}
\label{sect:method_general_importance}
In our graph (\Cref{fig:graph_main}), each node represents a sum of incoming vectors (edges). 
According to ALTI, the \textit{importance} of each vector (edge) to the overall sum (node) is \textit{proportional to its proximity} to the resulting sum. Formally, if $\vy = \vz_1 + \dots + \vz_m$,
%\vspace{-3pt}
\begin{equation}
    \!\!\!\!importance(\vz_j,\vy)\!=\! \frac{proximity(\vz_j, \vy)}{\sum\limits_{k}proximity(\vz_k, \vy)},
    \label{eq:contribution_main}
\end{equation}
\vspace{-5pt}
\begin{equation*}
    proximity(\vz_j,\vy) = \max(-||\vz_j - \vy||_1 + ||\vy||_1, 0).
    \label{eq:proximity}
\end{equation*}
%\vspace{-5pt}
Here, we use negative distance as a measure of similarity:
the smaller the distance between $\vz_j$ and~$\vy$, the more the information of~$\vz_j$ in~$\vy$. Note also that we ignore contributions of the vectors lying beyond the~$l_1$ length of~$\vy$. For more details, see~\citet{ferrando-etal-2022-measuring}.

Now, we need to define vector updates corresponding to the edges for the FFN and the attention blocks.

\subsubsection{Defining FFN Edges}
\label{sec:ffn_edges}
For the FFN blocks, edge vectors are straightforward. Following the notation of Figure~\ref{fig:graph_main}:
 \begin{align*}
 \label{eq:MLP_update}
    \vx_{pos}^{l} &= \textcolor{gray}{\vx_{pos}^{lA}} + \textcolor{violet}{\text{FFN}_{l}(\vx_{pos}^{lA})},
\end{align*}
where the terms correspond to the edges of the residual connection and FFN, respectively.

\begin{figure}[!t]
	\begin{centering}
	\includegraphics[width=0.48\textwidth]{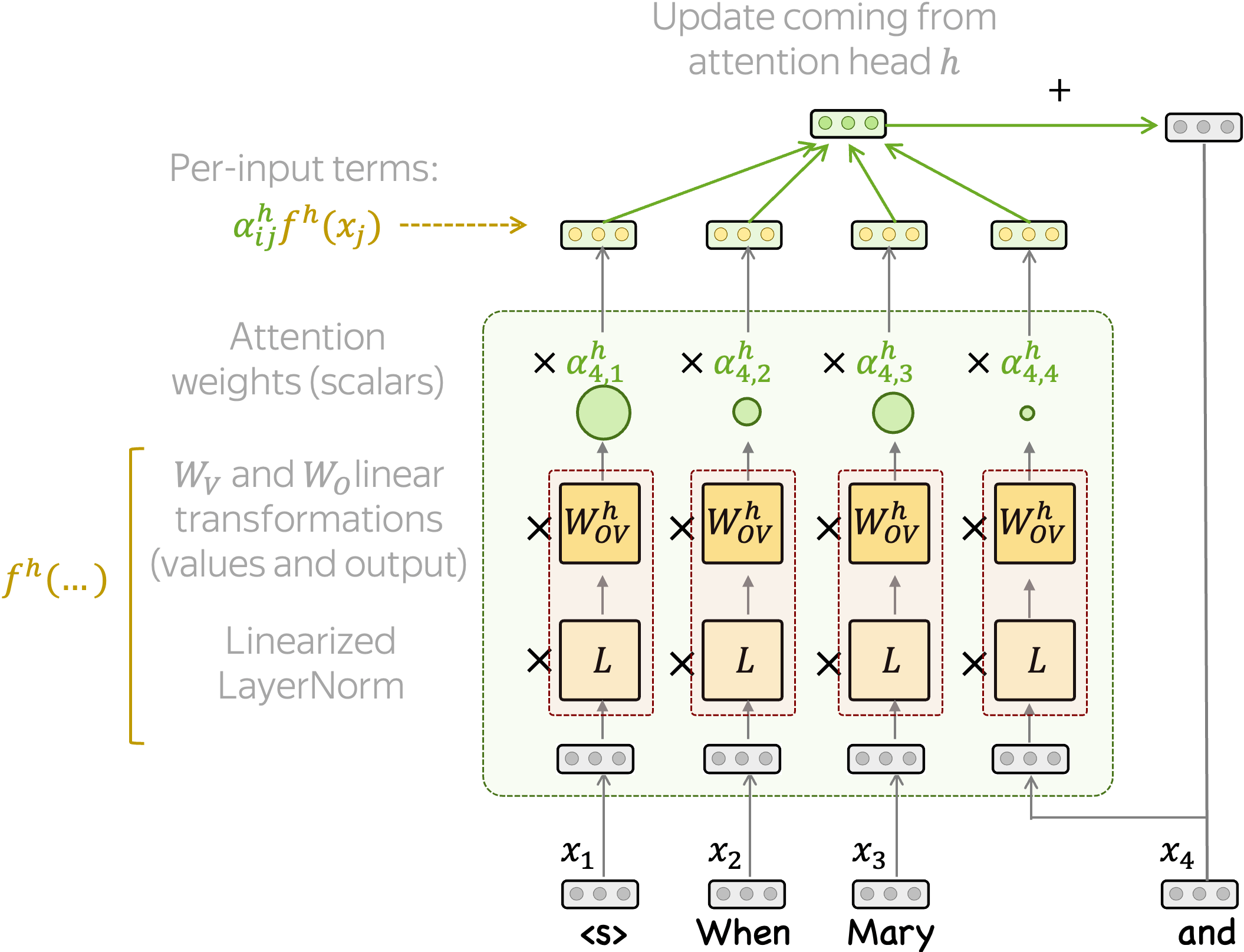}
	\caption{Decomposition of an update coming from an attention head into per-input terms. Layer indices are omitted for readability.}
	\label{fig:heads_output_mid_res}
	\end{centering}
\end{figure}

\subsubsection{Defining Attention Edges}
\label{sec:attn_edges}

We follow previous work~\citep{kobayashi-etal-2020-attention,ferrando-etal-2022-measuring} and decompose the output of an attention block into a sum of vectors (edges), each corresponding to a connection between residual streams~(\Cref{fig:heads_output_mid_res}).

\paragraph{Attention heads.} Formally, for attention head $h$
\begin{equation}\label{eq:decomposition_attn_head}
Attn^{h}(\mathbf{x}_{\le pos}) =  \sum_{j\le pos} \alpha_{pos,j}^{h} f^{h}(\vx_{j}),
\end{equation}
where $f^{h}(\vx_{j}) = \vx_jLW^{h}_{OV}$,  $W_{OV}^{h}=W_V^hW_O^h$ the values and output combined matrix for head $h$,
$\alpha_{pos,j}^{h}$ are scalar attention weights, and $L$ is the linearized layer normalization (see Appendix \ref{appx:linearizing_ln}).\footnote{From here onwards, we omit layer indices for readability.} 

Expanding further, a typical attention implementation (1)~weights corresponding value vectors within each attention head, (2)~concatenates head outputs, and (3)~further multiplies by the output matrix $W_O$. We split the output matrix $W_O$ into its head-specific parts $W_O^h$, drag these parts inside attention heads, and combine them with the head's values matrix: $W_{OV}^{h}=W_V^{h}W_O^{h}$.

Overall, Figure~\ref{fig:heads_output_mid_res}  shows that along an attention head each representation $\vx_j$ is transformed linearly into ${f^h(\vx_j)=\vx_jLW_{OV}^h}$ and multiplied by a scalar attention weight $\alpha_{pos,j}^{h}$. 
This gives us the ``raw output'' emitted by each input vector $\vx_j$ when treating attention weights as prediction-specific constants.
In this view, \textit{information flows through attention heads by independent channels $\alpha_{pos,j}^{h} f^{h}(\vx_{j})$ that converge in the next residual stream state}, we refer to each of this channels as a sub-edge.

\begin{figure}[!t]
	\begin{centering}
	\includegraphics[width=0.48\textwidth]{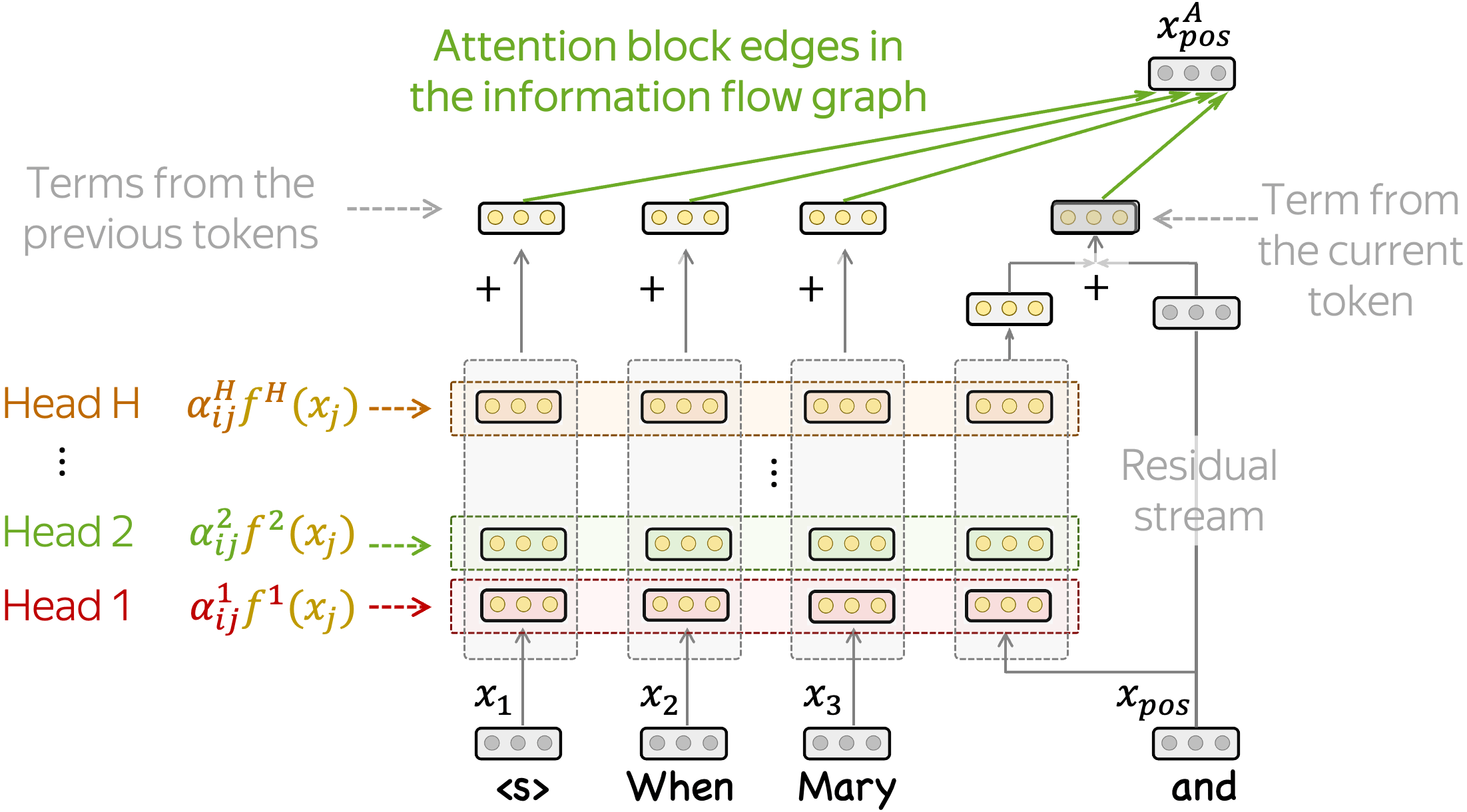}
	\caption{Decomposition of an update coming from an entire attention layer into per-input terms. Layer indices are omitted for readability.}
	\label{fig:heads_put_together}
	\end{centering}
\end{figure}

\paragraph{Attention block.} The information in an attention block flows through all the independent channels (sub-edges) in the $H$ heads~(\Cref{fig:heads_put_together}):
\begin{equation}\label{eq:decomposition_attn}
Attn(\mathbf{x}_{\le pos}) =  \sum\limits_{h}^H\sum_{j\le pos} \alpha_{pos,j}^{h} f^{h}(\vx_{j}).
\end{equation}
We compute the importance of each of the sub-edges in this sum as described in~\Cref{sect:method_general_importance} and aggregate across heads the capacities of those sub-edges connecting the same pair of nodes, $\sum_h^H e^{h}_{pos,j}$. Additionally, we include the importance of the residual connection for the current token, $e_{pos}^{\text{res\_attn}}$~(Figure~\ref{fig:heads_put_together}). Formally, attention edge importances are computed as
\vspace{-4.5pt}
\begin{equation}\label{eq:aggregating_head_scores}
\resizebox{0.35\textwidth}{!}{$\displaystyle{
  e^{\text{attn}}_{pos,j} = \left\{
  \begin{array}{@{}ll@{}}
    \sum_h^H e^{h}_{pos,j} & \mbox{if}~ j \neq pos \\
    \sum_h^H e^{h}_{pos,j} + e_{pos}^{\text{res\_attn}} & \mbox{if}~ j=pos
  \end{array}\right.}$}
  \vspace{-4.5pt}
\end{equation}
where ${e^{h}_{pos,j} = importance(\alpha_{pos,j}^{h} f^{h}(\vx_{j}), \vx^A_{pos})}$ and ${e_{pos}^{\text{res\_attn}} = importance(\vx_{pos}, \vx^A_{pos})}$ respectively, as defined in equation~(\ref{eq:contribution_main}).

\subsection{Extracting the Important Subgraph}
In~\Cref{sect:edge_importance_full} we explained how we compute the importances of the edges in the full information flow graph (Figure~\ref{fig:graph_main}). Finally, when building the important information flow subgraph (routes), we add only the edges with an importance above the specified threshold $\tau$~(\Cref{fig:algorithm}). Although we didn't notice significant differences, in our experiments we first remove the sub-edges with importances $e^{h}_{pos,j}$ and $e^{\text{res\_attn}}_{pos,j}$ below $\tau$, and renormalize the rest before aggregating across heads.
%One can make this algorithm more fine-grained by taking into account a few considerations.

Generally, feed-forward blocks have higher importance than attention heads. This is expected: while attention heads are only a (small) part of the attention block, the feed-forward block is not decomposed further. Therefore, one can set different thresholds for retaining attention and FFN edges, although we did not experiment with this.

\begin{figure*}[!t]
	\begin{centering}
	\includegraphics[width=0.95\textwidth]{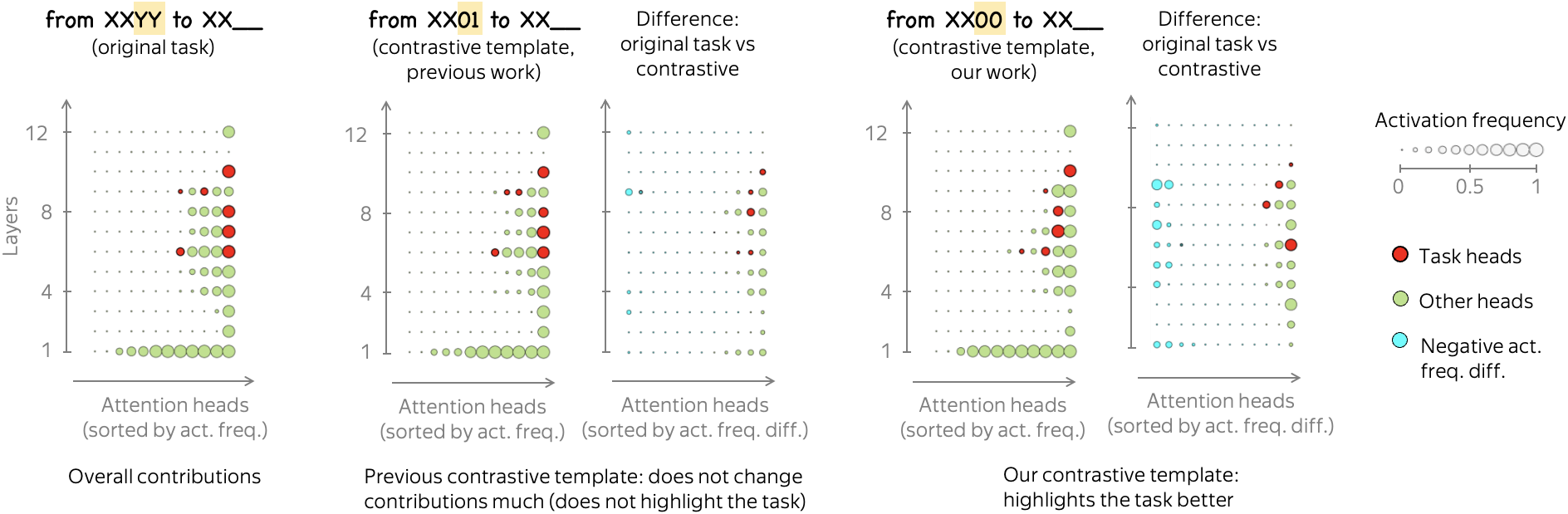}
	\caption{Greater-than, GPT2-Small. Attention head activation frequency ($\tau = 0.03$).}
	\label{fig:gpt2_small_greater_than_heads}
	\end{centering}
\end{figure*}

\begin{figure}[!t]
	\begin{centering}
	\includegraphics[width=0.48\textwidth]{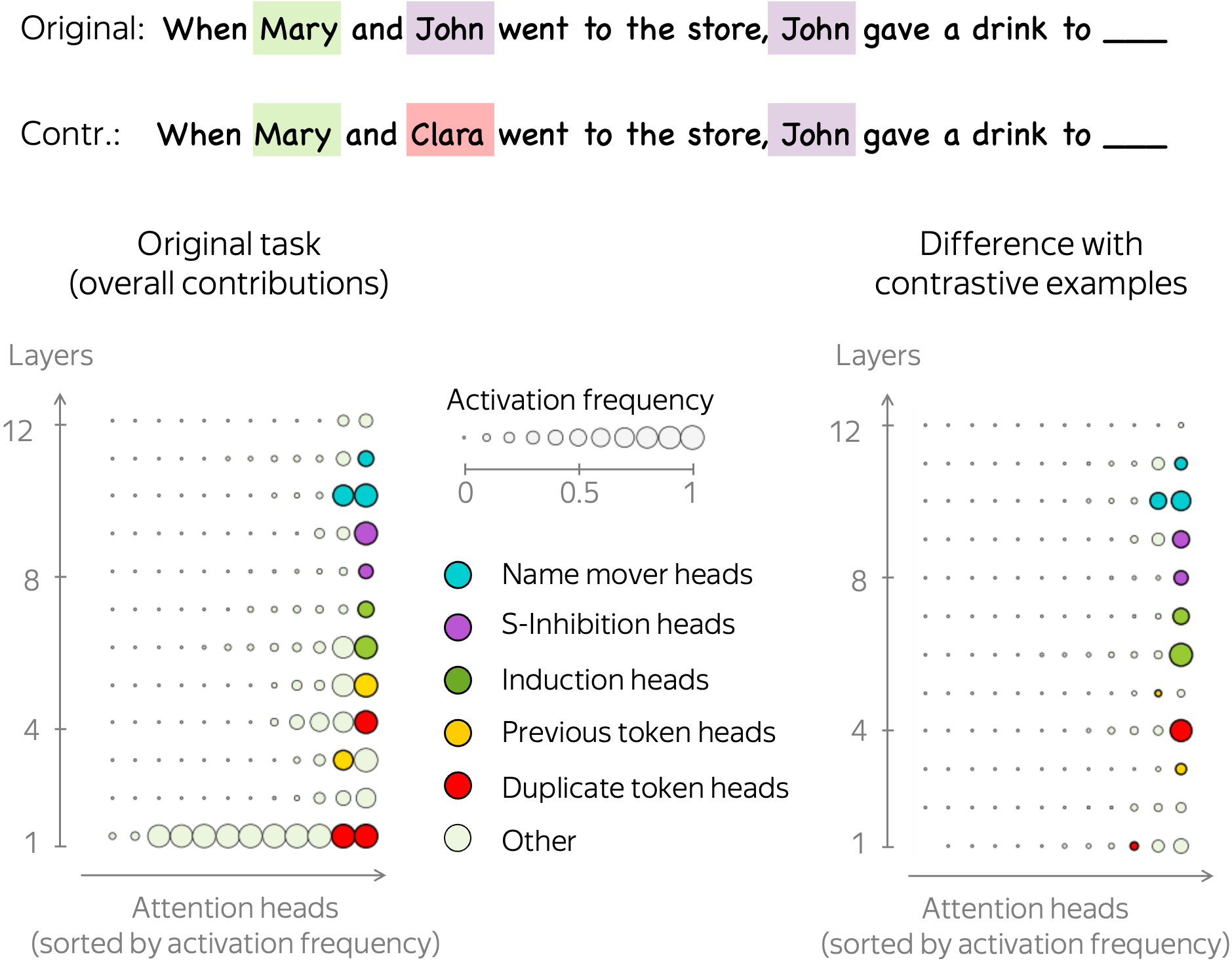}
	\caption{IOI, GPT2-Small. Attention head activation frequency ($\tau = 0.03$).}
	\label{fig:gpt2-small_ioi_correct_circuit_type_heads}
	\end{centering}
\end{figure}

\section{Information Flow vs Patching Circuits}

First, we compare our information routes to the circuits found in previous work for the Indirect Object Identification (IOI)~\citep{wang2023interpretability} and Greater-than~\citep{hanna2023does} tasks. The IOI task consists of predicting the next word in sentences like ``\textit{When Mary and John went to the store, John gave a drink to \_\_\_}''. An initial clause features the names of two individuals (Mary and John). Then, the second clause depicts a person exchanging an item with the other person. The goal is to predict the second name in the main clause, i.e. the Indirect Object~(Mary). In the Greater-than task, the model is prompted to predict a number given a sentence following the template: ``\textit{The [noun] lasted from the year XXYY to the year XX\_\_\_}''. Here, the task is to predict a number higher than YY.

\paragraph{Information flow vs patching.} 
The circuits for these tasks were previously found using activation patching with contrastive templates. These templates are designed to bring to the surface the specific task. For example, for IOI a contrastive template contains three different names instead of repeating one of them (Figure~\ref{fig:gpt2-small_ioi_correct_circuit_type_heads}). One of the differences between information flow and patching results is that our method finds all the components that contributed to the prediction, while patching finds what is important for the original task but not the contrastive baseline. As we will see, (i) our method can also be used in this manner, and (ii) it gives more reasonable results.

\subsection{Indirect Object Identification}
\label{sect:ioi}
For the IOI task, the previously found circuit contains several attention heads with various functions, e.g. Name Mover Heads, Duplicate Token Heads, etc.~\cite{wang2023interpretability}. Using our method, we extract information flow routes for both the original task and the contrastive templates, and evaluate the activation frequency of attention heads\footnote{We consider an attention head activated if it has at least a sub-edge in the information flow routes (important subgraph).}. When looking at overall contributions, Figure~\ref{fig:gpt2-small_ioi_correct_circuit_type_heads} (left) shows that our routes largely consist of the heads discovered previously. There are, however, several heads which are always part of the routes (have near 1 activation frequency) but are not in the ``patching circuit'' (shown with pale circles). Interestingly, when looking at the difference with contrastive templates (Figure~\ref{fig:gpt2-small_ioi_correct_circuit_type_heads}, right), these generic heads disappear. This makes sense: information flow routes contain all the components that were important for prediction, and these contain (i)~generic components that are overall important, and (ii)~task-specific heads. Note also how our method goes \textit{further than patching} for identifying the difference between the original and the contrastive tasks. Figure~\ref{fig:gpt2-small_ioi_correct_circuit_type_heads} shows that e.g. the previous token heads (yellow) are important for both types of predictions and are not specific to the IOI task. 

\subsection{Greater-than}
For the overall contributions in the greater-than task (Figure~\ref{fig:gpt2_small_greater_than_heads}, left), we also see that (i)~task-specific heads discovered via patching are among the most important heads in the information flow routes, and (ii)~many other heads are important. Let us now look at the differences with the contrastive templates.

\paragraph{Contrastive template matters.}
Interestingly, when using the same template as \citet{hanna2023does}, i.e. ``\textit{...from the year XXYY}'' with $\text{YY}=01$, instead of $\text{YY}>01$ overall contributions do not change much: all heads important for the original task are also important for the contrastive template (Figure~\ref{fig:gpt2_small_greater_than_heads}, center). While this contradicts the original work by \citet{hanna2023does}, this is expected: predicting a number higher than ``\textit{01}'' still requires greater-than reasoning (not all numbers would fit after ``\textit{01}'', e.g. ``\textit{00}'' would not). In contrast, if we consider another template with $\text{YY}=00$, overall contributions change and ``greater-than heads'' (red in Figure~\ref{fig:gpt2_small_greater_than_heads}) become less important. This difference in the results for ``\textit{01}'' vs ``\textit{00}'' in the contrastive template highlights the fragility of patching: not only it requires human-defined contrastive templates, but also  \textit{patching results are subjective since they vary depending on the chosen template}.

Overall, we see that, compared to patching, information flow routes are more versatile and informative. Indeed, they can find the importance of model components both (i)~overall for a prediction, and (ii)~compared to a contrastive example (i.e., patching setting), and are able to show differences between the contrastive templates.

\subsection{OPT-125m vs GPT2-Small}

Additionally, we experiment with OPT-125m which has the same number of layers and heads than GPT2-small. We find that the information flow routes for IOI and greater-than tasks are similar for OPT-125M and GPT2-small (\Cref{sect_apdx:examples}).

\subsection{Hundred Times Faster than Patching}
Obtaining information flow routes involves a two-step process. First, we run a forward pass and cache internal activations. Then, we obtain edge importances to build the subgraph as depicted in Algorithm~\ref{fig:algorithm}. For comparison purposes, we contrast our approach with the ACDC algorithm~\citep{conmy2023automated}, an automated circuit discovery algorithm based on patching. According to their findings, on the IOI task with a batch of 50 examples, it requires 8 minutes on a single GPU to discover the circuit. In contrast, our method accomplishes this task in 5 seconds, around 100x in time reduction.

\begin{figure}[!t]
	\begin{centering}
	\includegraphics[width=0.45\textwidth]{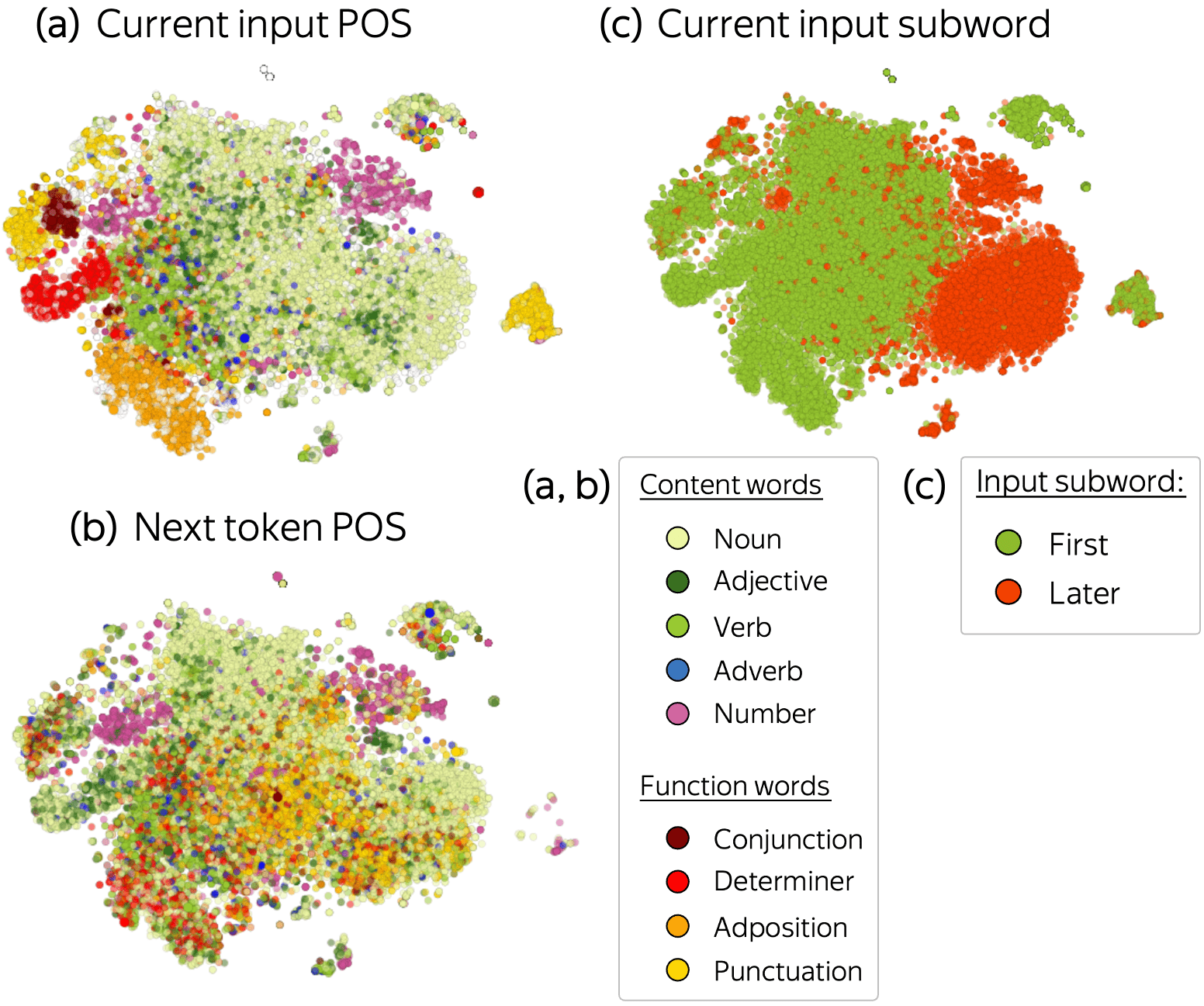}
	\caption{t-SNE of component importance vectors. Coloured by:  (a)~input token POS Tag, (b)~next token POS tag, (c)~whether the input token is the first or a later subword. Llama 2-7B.}
 \label{fig:tsne_c4_javi_pos_tag_univ}
	\end{centering}
\end{figure}

\section{General Experiments}
Information flows inside the network via mechanisms implemented in the model. In this section, we look at general patterns in component importances, and try to understand if important mechanisms depend on the type of tokens processed. Using the full information flow graph (Figure~\ref{fig:graph_main}) of Llama 2-7B, we look at all the immediate connections with each residual stream. We then record the importance values of all the sub-edges corresponding to individual attention heads (i.e., $e_{pos,j}^h$), as well as FFN blocks. We use a subset of 1000 sentences from the C4 dataset~\citep{c4_t5}.

\subsection{Component Importance for POS }
\label{sect:tsne_pos}

First, let us see how component importance depends on parts of speech (POS).
For this, we pack per-prediction component importances into vectors\footnote{Each vector corresponding to the $pos$-th position is defined as $\left(\sum_j e^{1,1}_{pos,j}, \sum_j e^{1,2}_{pos,j}, \ldots, \sum_j e^{L,H}_{pos,j}, e^{\mbox{ffn}_1}_{pos}, \ldots, e^{\mbox{ffn}_L}_{pos}\right)$.} and apply t-SNE~\citep{JMLR:v9:vandermaaten08a}. \Cref{fig:tsne_c4_javi_pos_tag_univ} shows the resulting projection with datapoints colored according to either input or next token\footnote{Here, we take the next token from the dataset and not the one generated by the model. While this adds some noise to the results, we expect at least parts of speech of the reference and the predicted tokens to be similar.} part of speech.\footnote{We obtain POS tags with NLTK (universal tagset,~\citet{petrov-etal-2012-universal}) and assign a tag to all subwords of each word.}

\paragraph{Content vs function words.}
From \Cref{fig:tsne_c4_javi_pos_tag_univ}a we see that for function words as inputs, component contributions are clustered according to their part of speech tag. Roughly speaking, for these parts of speech the model has ``typical'' information flow routes and, by knowing component contributions, we can infer input token POS rather accurately. Interestingly, this is not the case for content words: while we can see verbs being somewhat separated from nouns, overall contribution patterns for content words are mixed together in a large cluster. Apparently, the reasoning for these words is more complicated and is defined by a broader context rather than simply input token POS.

\paragraph{First vs later subwords.} 
Diving deeper, \Cref{fig:tsne_c4_javi_pos_tag_univ}c shows that contribution patterns strongly depend on whether the current token is the first or a later subword of some word. Clearly, there are some model components specialized for first or later subwords~-- we confirm this later in Section~\ref{sect:positional_subword_heads}. Note also that \Cref{fig:tsne_c4_javi_pos_tag_univ}c explains the two distinct clusters for numbers we saw in \Cref{fig:tsne_c4_javi_pos_tag_univ}a (purple): they separate number tokens into first and later digits.

\paragraph{Patterns wrt to current vs next token.} Finally, \Cref{fig:tsne_c4_javi_pos_tag_univ}b shows the same datapoint colored by the part of speech of the next token. Comparing this to  \Cref{fig:tsne_c4_javi_pos_tag_univ}a, we see that contribution patterns depend more on input tokens rather than output tokens. This might be because in the lower part of the network, a model processes inputs in a generic manner, while the higher network part (that prepares to generate output) is more fine-grained and depends on more attributes than part of speech.

\subsection{Bottom-to-Top Patterns}
\label{sect:bottom_to_top}

\begin{figure}[t!]
	\begin{centering}
	\includegraphics[width=0.48\textwidth]{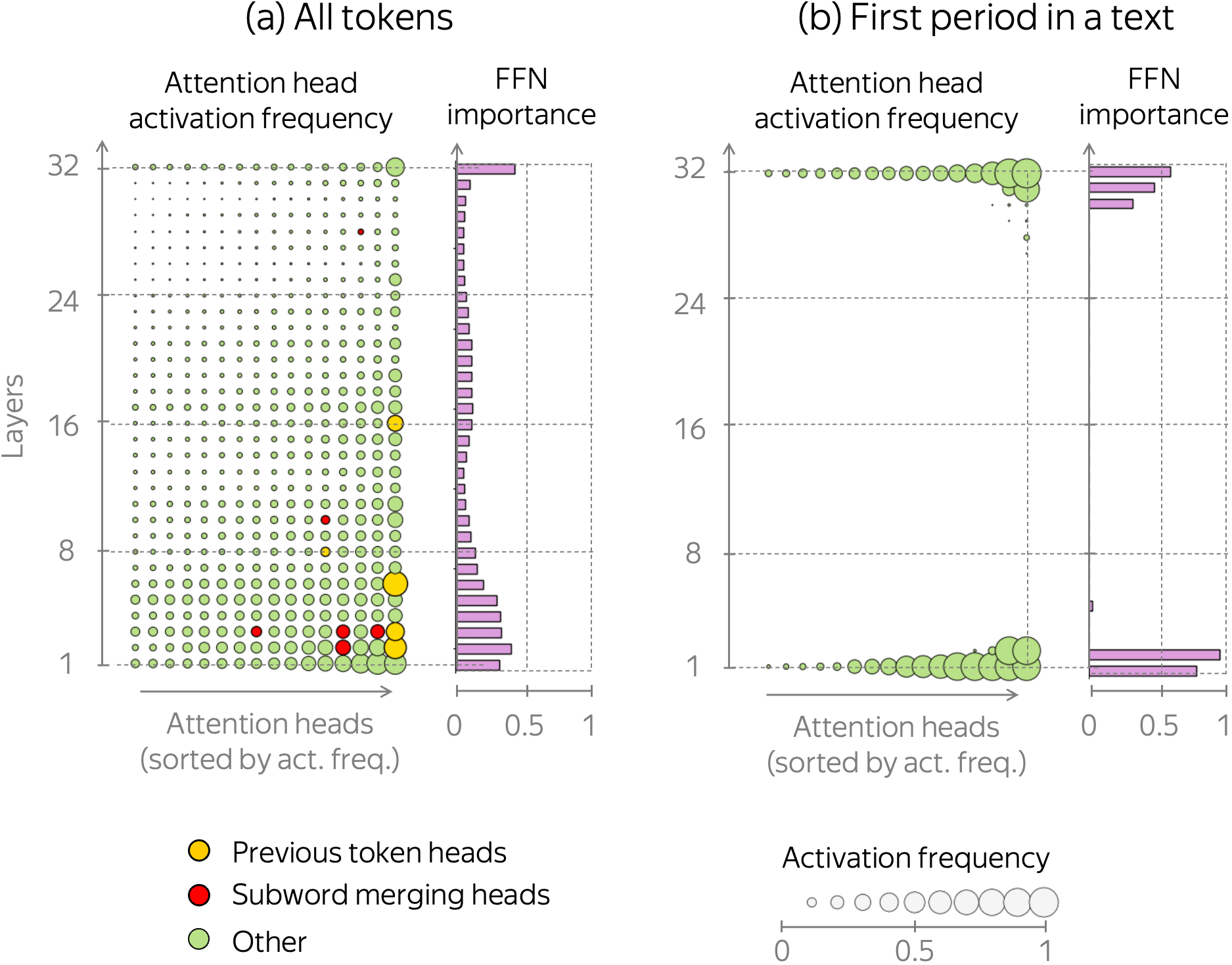}
	\caption{Attention head activation frequency ($\tau=0.01$) and FFN block importance. We show only top-50$\%$ important heads. Llama 2-7B.}
 \label{fig:bottom_to_top}
	\end{centering}
\end{figure}

Since we already started talking about functions of the lower and the higher parts of the network, let us now look at the bottom-up contribution patterns. Figure~\ref{fig:bottom_to_top}a shows 
as average activation frequency of attention heads for a small $\tau=0.01$~-- this gives us an estimate ``of'' how many times an attention head affects a residual stream. Additionally, we show the importance of the FFN block in each layer. As expected, attention and feed-forward blocks are more active at the bottom part of the network, where the model processes input and extracts general information. Later, when a model performs more fine-grained and specialized operations, the activation frequency of individual attention heads and FFN blocks goes down. Interestingly, the last-layer FFN is highly important: apparently, this last operation between the residual stream and the unembedding matrix (i.e., prediction) changes representation rather significantly.

\subsection{Positional and Subword Merging Heads}
\label{sect:positional_subword_heads}
Some of our observations above hint at certain attention head functions that might be overall important. First, in Section~\ref{sect:ioi} we showed that previous tokens heads are generally important in the IOI task, both for target examples and the contrastive baseline~-- this attention head function might be important in general. In Section~\ref{sect:tsne_pos} we noticed a clear difference between component contributions for first and later subwords~-- this suggests that some model components might be responsible for that.

In what follows, we look at two attention head functions found earlier for machine translation, positional and subword merging heads, and see whether they are generally important. Differently from previous work, we define a head function based on token contributions within the head and not attention weights as done previously, since attention weights might not reflect influences properly~(\citealp{bastings-filippova-2020-elephant,kobayashi-etal-2020-attention}, among others).

\paragraph{Positional heads.} Originally, previous token heads were found to be the most important attention heads in machine translation encoders~\cite{voita-etal-2019-analyzing}. Now, let us check their importance for LLMs.  We refer to a head as the previous token head if in at least 70$\%$ of the cases it puts more than half of its influence on the previous token.

Figure~\ref{fig:bottom_to_top}a shows previous token heads in yellow. As in the earlier work for machine translation~\cite{voita-etal-2019-analyzing}, we also see that for LLMs, (i)~there are several previous tokens heads in the model, and (ii)~almost all of them are by far the most important attention heads in the corresponding layers.

\paragraph{Subword merging heads.} Putting together our first-vs-later subword observations in Section~\ref{sect:tsne_pos} and previous work, we might expect our model to have subword merging heads found for machine translation encoders~\cite{correia-etal-2019-adaptively}.
We refer to a subword merging head if later subwords take information from previous subwords of the same word but the head is \textit{not} previous token head (see~\Cref{apx:subword_merging} for a formal definition). We see subword merging heads in the bottom part of the network~(Figure~\ref{fig:bottom_to_top}a, red), and are among the most important heads in the corresponding layers. Notably, previous work did not study their general relevance~\cite{correia-etal-2019-adaptively}. Note that these subword merging heads are important for later subwords and not important otherwise~-- this explains the clusters we saw in Figure~\ref{fig:tsne_c4_javi_pos_tag_univ}c. 

We would like to highlight that for LMs, we are the first to talk about \textit{general importance of attention heads} and to find that some of the most important heads are previous token and subword merging heads. Future work might explore the functions of other important heads in the model.

\begin{figure}[t!]
	\begin{centering}
	\includegraphics[width=0.4\textwidth]{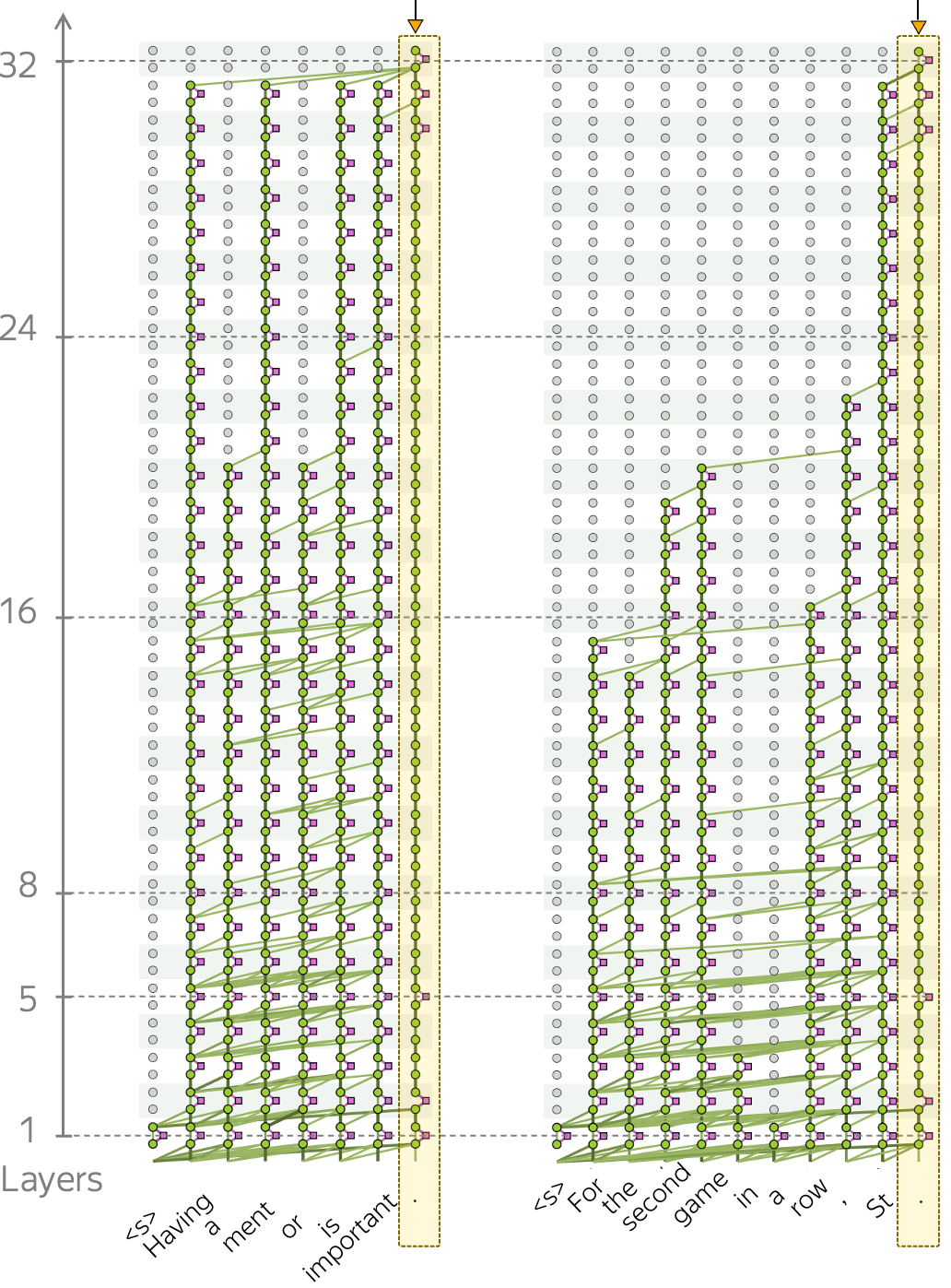}
	\caption{Examples of important information flow subgraphs ($\tau=0.01$). Llama 2-7B.}
 \label{fig:special_examples}
	\end{centering}
\end{figure}

\subsection{Peculiar Information Flow Patterns, or Periods Acting as BOS}
\label{sect:peculiar_patterns}

In Section~\ref{sect:tsne_pos} we talked about general contribution patterns and saw visible clusters corresponding to the input tokens' part of speech. Now, let us go deeper and look in detail at one of the clusters. We choose the outlier punctuation cluster shown in yellow in
Figure~\ref{fig:tsne_c4_javi_pos_tag_univ} (to the right)~-- this cluster corresponds to the first period in a text.\footnote{823 out of 826 datapoints.}

\begin{figure*}[h!]
	\begin{centering}
	\includegraphics[width=0.85\textwidth]{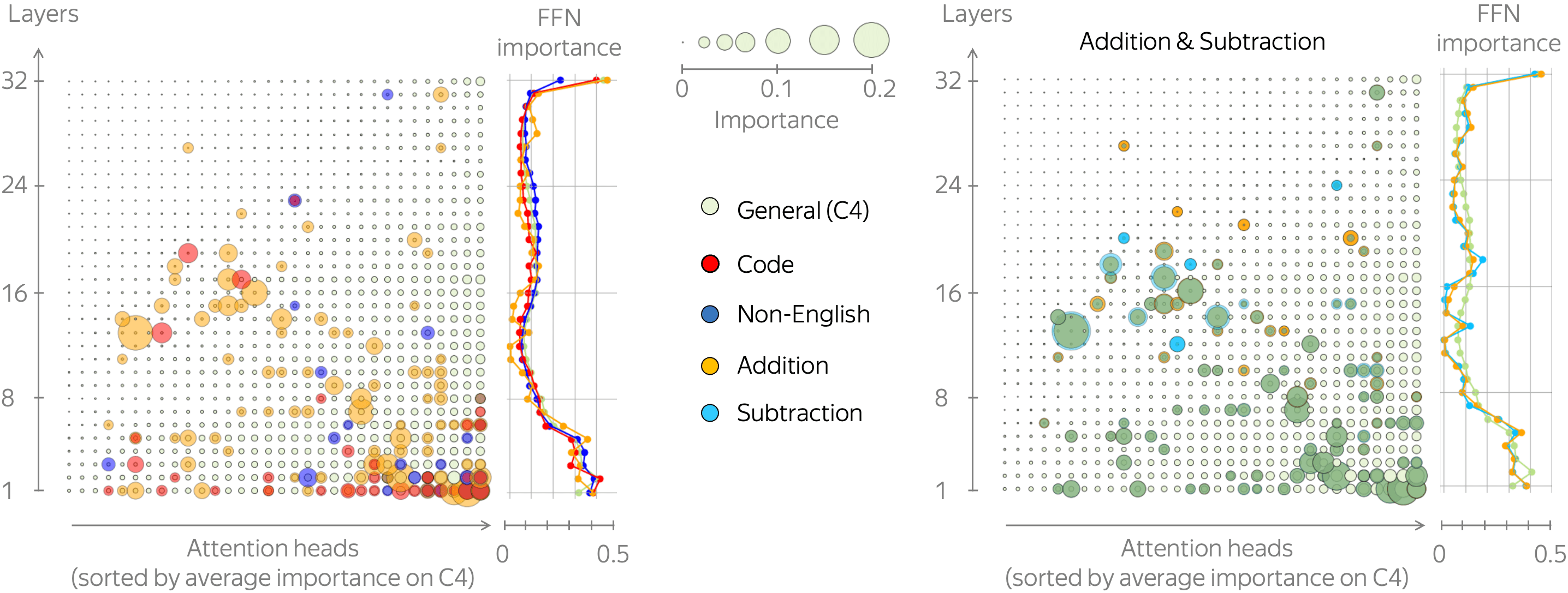}
	\caption{Average importance of attention heads and FFNs for different datasets. For non-general domains, we show only heads with importance higher than 0.015. Llama 2-7B.}
 \label{fig:specialization}
	\end{centering}
\end{figure*}

Figure~\ref{fig:bottom_to_top}b shows the average importance of model components for examples in this cluster. We see that for these examples, the residual stream ignores all attention and FFN blocks in all the layers except for the first and last few: for most of the layers, contributions of all model components are near zero. 
When we look at the information flow graphs for these examples, we see that, even for a rather small threshold $\tau=0.01$, bottom-to-top processing happens largely via ``untouched'' residual connection (Figure~\ref{fig:special_examples}). In Appendix~\ref{sect_apdx:period_bos} we show that up to the last three layers, this residual stream \textit{takes the role of the BOS token} and future tokens treat this residual stream as such. Concurrent work~\citep{cancedda2024spectral} suggests that early FFN's update into a particular subspace is responsible for this behavior. While for some examples periods acting as BOS might be reasonable, this also happens in cases where the period does not have an end-of-sentence role. For example, Figure~\ref{fig:special_examples} (right): while in a sentence \texttt{For the second game in a row, St. Thomas ...} the first period does not have the end-of-sentence meaning, the residual stream is still acting in the same manner. In future work, it might be valuable to explore whether this behavior might cause incorrect generation behavior.

\section{Domain-Specific Model Components}

\begin{figure*}[!t]
	\begin{centering}
	\includegraphics[width=0.8\textwidth]{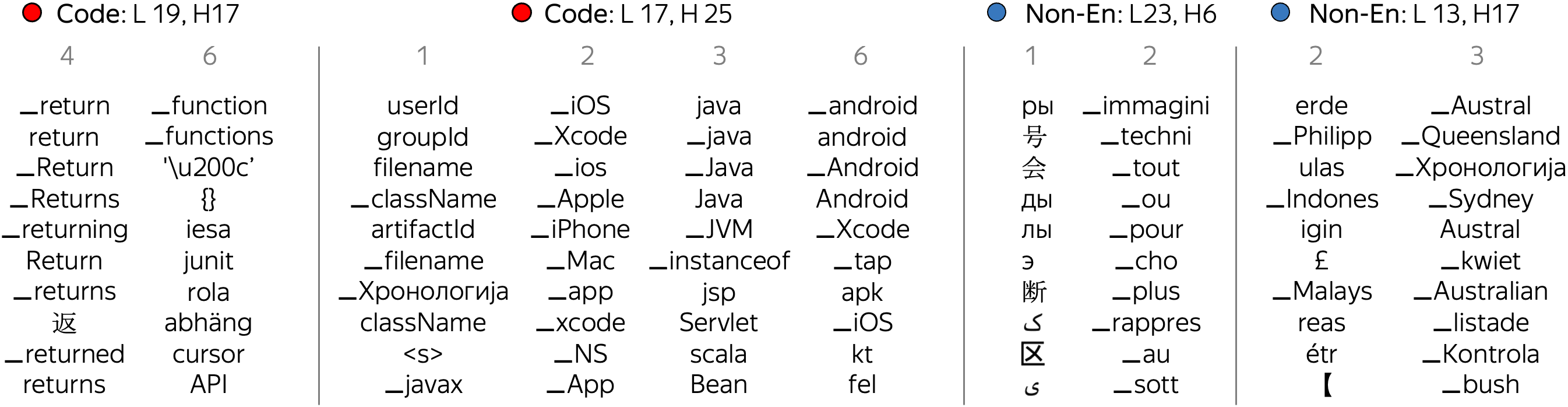}
	\caption{Top 10 tokens after projecting singular values of $W_{OV}$ onto the unembedding matrix. Singular value ids are shown in gray.}
	\label{fig:svd_lang_code}
	\end{centering}
 \vspace{-2ex}
\end{figure*}

We have seen that some attention heads' roles are generally important across predictions. Now, we analyze if some model components are instead specialized in specific domains. We evaluate the average importance of Llama 2-7B's components on 1000 sentences from different domains. Specifically, we consider (i)~C4~\cite{c4_t5}, (ii)~English, Russian, Italian, Spanish \textsc{FloRes-200} devtest sets~\citep{nllb2022,goyal-etal-2022-flores}, (iii)~code data from CodeParrot\footnote{\url{https://huggingface.co/datasets/codeparrot/github-code}}, (iv)~addition/subtraction data~\cite{stolfo2023understanding}.

\subsection{Components are Specialized}
Figure~\ref{fig:specialization} shows the importance of attention heads and FFN blocks in general and specific datasets.

\paragraph{Coarse grained: domains and languages.} We see that generally unimportant attention heads become highly relevant for specific tasks. For example, the most important heads for addition are among the lowest scoring heads when looking at C4 (Figure~\ref{fig:specialization}~(left), large yellow blobs to the left). Another interesting observation is that domain-specific heads are different across domains: important heads for addition, code, and non-English are not the same heads. Overall, we can see that model components are largely specialized.

\paragraph{Fine-grained: addition vs subtraction.}  Figure~\ref{fig:specialization}~(right) shows more fine-grained results where instead of different domains, we look at tasks within the same narrow domain: addition vs subtraction. While the important heads for addition and subtraction largely intersect, we see several heads that are active only for one task and not the other (bright blue and yellow blobs). This suggests that this fine-grained specialization might be responsible for ``reasoning'' inside the model and not just domain-specific processing; future work may validate this further. Interestingly, we did not find similar specialization between languages and only find non-English heads~-- probably, portions of training data in these languages were not large enough to have dedicated language-specific heads.

Finally, we can also see the difference in the importance of the feed-forward blocks with respect to each domain. For example, the last FFN layer is much less relevant for non-English than other domains (Figure~\ref{fig:specialization}, left), and the importance of FFN blocks for addition and subtraction falls to zero in some layers (Figure~\ref{fig:specialization}, right). In future work, one might conduct a more fine-grained analysis of the importance of feed-forward layers by looking at the relevance and functions of individual neurons~(along the lines of \citep{geva-etal-2021-transformer,dai-etal-2022-knowledge,voita2023neurons} but taking into account neuron importance for specific predictions).

\subsection{Specialized Heads Output Topic-Related Concepts}
\label{sec:svd}

In this section, we analyze what our domain-specific heads write into the residual stream, and find that some of them write highly interpretable and topic-related concepts.

\paragraph{Weight matrices analysis with SVD.}
As we illustrated in Figure~\ref{fig:heads_output_mid_res}, $W^{h}_{OV}$ transforms representations from each of the residual streams\footnote{After applying layer normalization first.} into vectors that are added to the current residual stream. To understand what kind of information is embedded in this transformation, we use Singular Value Decomposition (SVD). To get an intuitive explanation of the $W^{h}_{OV}$ impact, we can factorize it via the ``thin'' singular value decomposition~\citep{svd_weights} as $W^{h}_{OV} = U \Sigma V^{T}$.\footnote{$U \in \mathbb{R}^{d \times r}$, $\Sigma \in \mathbb{R}^{r \times r}$, $V^{T} \in \mathbb{R}^{r \times d}$; $d$ is vector dimensionality in the residual stream, $r$ is the rank of $W^{h}_{OV}$.} Then, projecting $\vx \in \mathbb{R}^{1 \times d}$ through $W^{h}_{OV}$ can be expressed as
\vspace{-9pt}
\begin{equation}
    \vx W^{h}_{OV} = (\vx U \Sigma) V^{T}
    = \sum_{i=1}^{r} (\vx \vu_i\sigma_i) \vv_i^T.
\end{equation}
\vspace{-10pt}

Here, each $\vu_i\sigma_i \in \mathbb{R}^{d \times 1}$ can be interpreted as a key that is compared to the query ($\vx$) via dot product~\citep{molina2023traveling}. Each query-key dot-product weights the right singular vector $\vv_i^T$. If we project these right singular vectors to the unembedding matrix ($\vv_i^T W_{U}$), we get an interpretation of the attention head's influence in terms of concepts (i.e., tokens) it promotes in the residual stream.

\paragraph{Top singular values.}
For some of the heads specific to code and non-English inputs we saw in~\Cref{fig:specialization}, 
in~\Cref{fig:svd_lang_code} we show the top 10 tokens that come from the described above projection. We see that code-specific heads promote tokens related to coding and technology in a more general sense: e.g.,  tokens related to Apple, like iOS, Xcode, iPhone, etc. Heads most active for non-English promote tokens in multiple languages, avoiding English ones. Also, we see tokens related to cities (Sydney), countries (Indones), and currencies (£). Overall, we looked at the functions of attention heads from two perspectives: (i)~when a head is active, and (ii)~how it updates the residual stream, and found they are consistent. While a similar kind of analysis was done before for neurons~\cite{voita2023neurons}, our method made it possible to talk about entire model components being active/non-active for a prediction.

\section{Related Work}
Earlier works evaluating the importance of model components include~\citet{bau2018identifying}, who identified important neurons in NMT models shared across models, and \citet{voita-etal-2019-analyzing} who looked at the relevance of entire attention heads in the Transformer from two perspectives: attribution and pruning. More recent research focuses more on task-specific rather than overall importance. They aim to find subparts of LMs responsible for these tasks and rely on activation patching methodology, firstly introduced to analyze biases in LMs by~\citet{causal_mediation_bias}, and used across several others~\citep{docstring}. Initial works discovered circuits for different tasks of GPT2, like IOI \citep{wang2023interpretability} and greater-than~\citep{hanna2023does}. Patching has been also used to locate factual knowledge in LMs~\citep{meng2022locating}, and to discover task vectors behind in-context learning capabilities~\citep{hendel2023incontext, todd2023function}.
Our work can be related to the concurrent line of research developing methods to approximate activation patching~\citep{syed2023attribution,kramár2024atp,hanna2024faith}.

\section{Conclusions}

We view computations inside the Transformer as information flowing between token representations through model components. Using this view, we propose to interpret language model predictions by extracting the important part of the overall information flow.
Our method for extracting these important information flow routes is automatic, highly efficient, applicable to any prediction,
more versatile and informative compared to existing pipelines.

\section{Acknowledgments}
We would like to thank Yihong Chen, Christoforos Nalmpantis, Igor Tufanov, Nicola Cancedda and Eduardo Sánchez for the insightful discussions.

\bibliography{custom}
\bibliographystyle{icml2024}

%%%%%%%%%%%%%%%%%%%%%%%%%%%%%%%%%%%%%%%%%%%%%%%%%%%%%%%%%%%%%%%%%%%%%%%%%%%%%%%
%%%%%%%%%%%%%%%%%%%%%%%%%%%%%%%%%%%%%%%%%%%%%%%%%%%%%%%%%%%%%%%%%%%%%%%%%%%%%%%
% APPENDIX
%%%%%%%%%%%%%%%%%%%%%%%%%%%%%%%%%%%%%%%%%%%%%%%%%%%%%%%%%%%%%%%%%%%%%%%%%%%%%%%
%%%%%%%%%%%%%%%%%%%%%%%%%%%%%%%%%%%%%%%%%%%%%%%%%%%%%%%%%%%%%%%%%%%%%%%%%%%%%%%
\newpage
\appendix
\onecolumn

\section{Background on Activation Patching}\label{apx:background_patching}
Activation patching~\citep{causal_mediation_bias,meng2022locating,geiger-etal-2020-neural,wang2023interpretability} refers to intervening some internal activation (intermediate representation) computed by a model component $c$ (attention head, feedforward network) in the forward pass ($f^c(x)$) with ‘base’ input $x$. The patched activation is taken from a forward pass $f^c(\tilde{x})$ on a ‘source’ input $\tilde{x}$. We can express this intervention using the do-operator~\citep{pearl_2009} as $f(x|\text{do}(f^c(x)=f^c(\tilde{x})))$. Upon intervention, the forward pass continues and the model output is compared with the prediction with the ‘base’ input, e.g. by measuring $f(x)-f(x|\text{do}(f^c(x)=f^c(\tilde{x})))$.

Identifying subnetworks (circuits) through activation patching has several shortcomings:
\begin{itemize}
\item It requires large human efforts to create the input base templates ($x$) and contrastive source examples ($\tilde{x}$) for the specific task to study, thus results vary depending on the choice of the contrastive template. Additionally, analyses are constrained to those templates, preventing from studying models on more general types of predictions.

\item For each prediction, one needs to patch every edge (or node) in the computational graph, which becomes impractical when studying large language models.

\item It has been shown that downstream components can compensate for the ablation as a form of self-repair~\citep{mcgrath2023hydra,rushing2024explorations}, which interferes with the analysis.
\end{itemize}
In contrast, our method doesn’t require specific templates with contrastive examples which allows us to study specific tasks and more general behaviors, computes the information flow routes graph in a single forward pass, and it’s not affected by self-repair issues, since we make no interventions.

\section{Details about the Information Flow Routes}
\subsection{Linearizing the Layer Normalization}\label{appx:linearizing_ln}

Given an input representation $\vx$, the layernorm computes
\begin{equation}
\text{LN}(\vx)=\frac{\vx-\mu(\vx)}{\sigma(\vx)} \odot \mathbf{\gamma}+ \mathbf{\beta}
\end{equation}
with $\mu$ and $\sigma$ obtaining the mean and standard deviation, and $\gamma \in \mathbb{R}^{d}$ and $\beta \in \mathbb{R}^{d}$ refer to learned element-wise transformation and bias respectively. Considering $\sigma(\vx)$ as a constant, $\text{LN}$ can be treated as a constant affine transformation:
\begin{equation}\label{eq:ln_linearization}
\text{LN}(\vx) = \vx L + \mathbf{\beta}
\end{equation}
where $L \in \mathbb{R}^{d \times d}$ represents a matrix that combines centering, normalizing, and scaling operations together.

\begin{equation*}
\resizebox{0.48\textwidth}{!}{$\displaystyle{
\mathbf{L}:=\frac{1}{\sigma(\vx)}\left[ 
\begin{array}{cccc}
\gamma _{1} & 0 & \cdots  & 0 \\ 
0 & \gamma _{2} & \cdots  & 0 \\ 
\cdots  & \cdots  & \cdots  & \cdots  \\ 
0 & 0 & \cdots  & \gamma _{n}%
\end{array}%
\right] \left[ 
\begin{array}{cccc}
\frac{n-1}{n} & -\frac{1}{n} & \cdots  & -\frac{1}{n} \\ 
-\frac{1}{n} & \frac{n-1}{n} & \cdots  & -\frac{1}{n} \\ 
\cdots  & \cdots  & \cdots  & \cdots  \\ 
-\frac{1}{n} & -\frac{1}{n} & \cdots  & \frac{n-1}{n}%
\end{array}%
\right]}$}
\end{equation*}%

The linear map on the right subtracts the mean to the input vector, $\vx' = \vx-\mu(\vx)$. The left matrix performs the hadamard product with the layer normalization weights ($\vx' \odot \gamma$).

\subsection{Subword Merging Head definition}\label{apx:subword_merging}
Formally, for a subword merging head (i)~in at least 70$\%$ of the cases, later subwords put more than half of their influence on previous subwords of the same word, (ii)~in at least 70$\%$ of the cases, for the first subword this head's overall influence is no more than $0.005\%$.

\begin{comment}
\subsection{Folding the Layernorm}
\label{appx:folding_ln}

Any Transformer block reads from the residual stream by normalizing before applying a linear layer (with weights $W$ and $\vb$) to the resulting vector:
\begin{equation}
    \text{LN}(\vx_j)W + \vb
\end{equation}

Following the reformulation of the layernorm shown in Eq. \ref{eq:ln_linearization}, we can fold the weights of the layernorm into those of the subsequent linear layer as follows:
\begin{align*}\label{eq:folded_ln}
\text{LN}(\vx_j)W + \vb &= \left(\frac{1}{\sigma(\vx_j)} \vx_j L + \beta \right)W + \vb\\
&= \frac{1}{\sigma(\vx_j)}\vx_jLW + \beta W + \vb\\
&= \frac{1}{\sigma(\vx_j)}\vx_jW^* + \vb^*\numberthis
\end{align*}
where $W^* = LW$ and $\vb^* = \beta W + \vb$.
\end{comment}

\clearpage

\section{Examples of Routes}
\label{sect_apdx:examples}

\begin{figure*}[!h]
	\begin{centering}
	\includegraphics[width=0.7\textwidth]{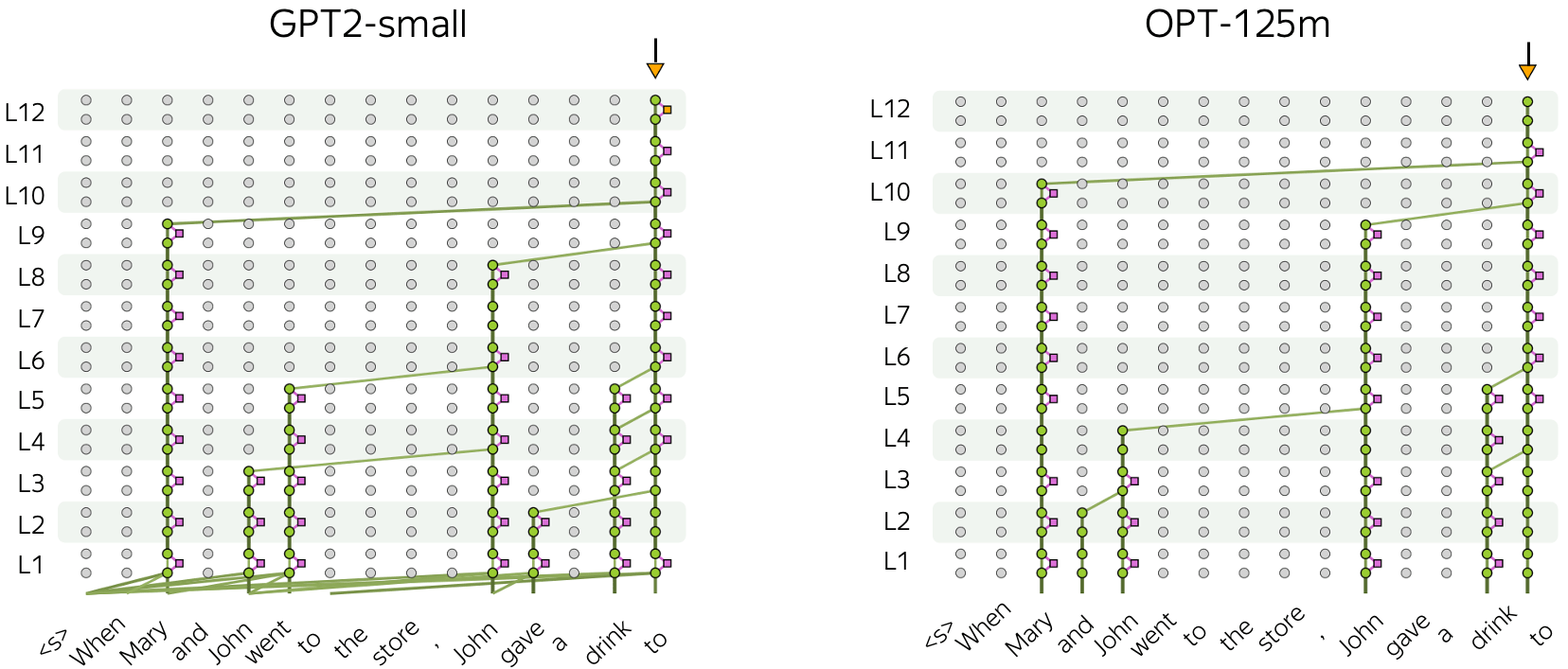}
	\caption{The important information flow routes, IOI task, $\tau=0.04$.}
	\label{fig:gpt_opt_ioi}
	\end{centering}
\end{figure*}

\begin{figure*}[!h]
	\begin{centering}
	\includegraphics[width=0.7\textwidth]{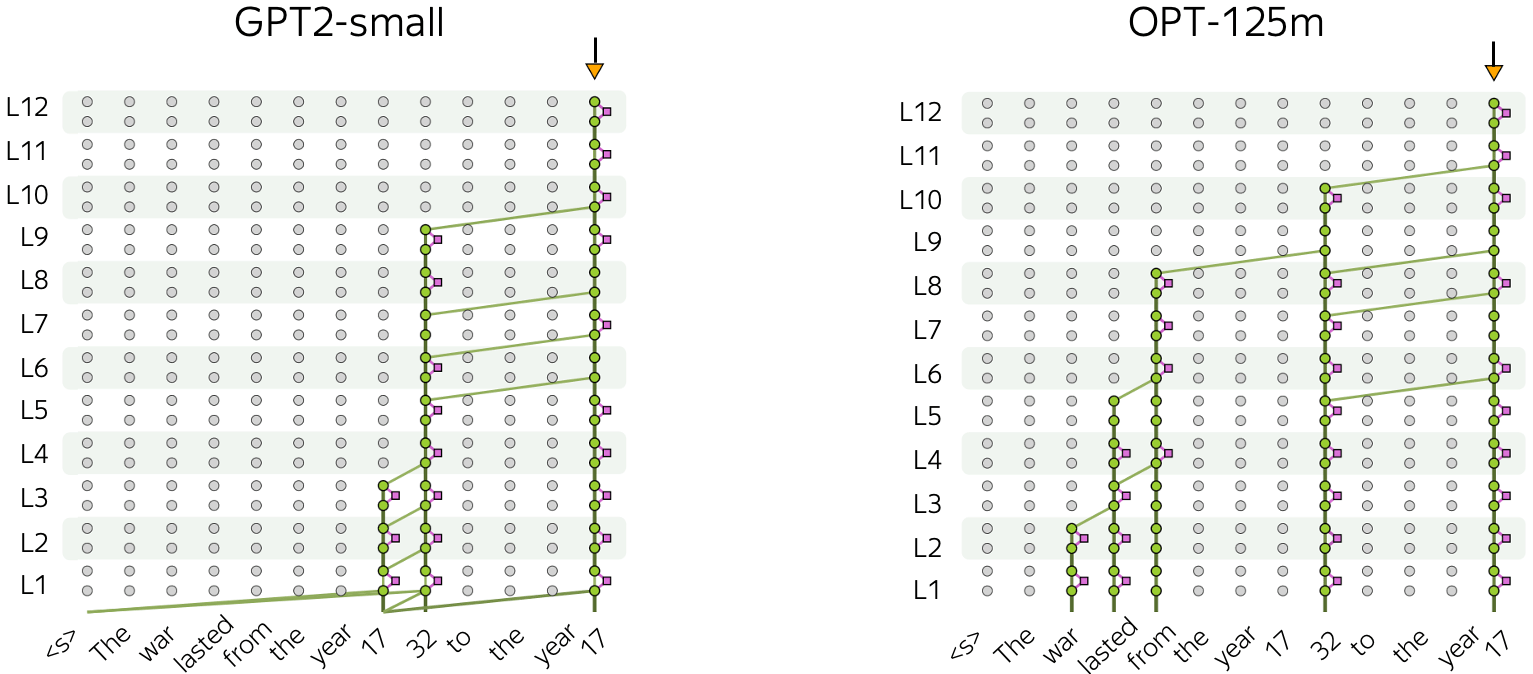}
	\caption{The important information flow routes, greater-than task, $\tau=0.04$.}
	\label{fig:gpt_opt_greater}
	\end{centering}
\end{figure*}

Figures~\ref{fig:gpt_opt_ioi} and \ref{fig:gpt_opt_greater} show the information flow routes for IOI and greater-than tasks extracted for GPT2-small and OPT-125m.

\clearpage
\section{Period Acting as BOS}
\label{sect_apdx:period_bos}

\begin{figure}[!h]
	\begin{centering}
	\includegraphics[width=0.46\textwidth]{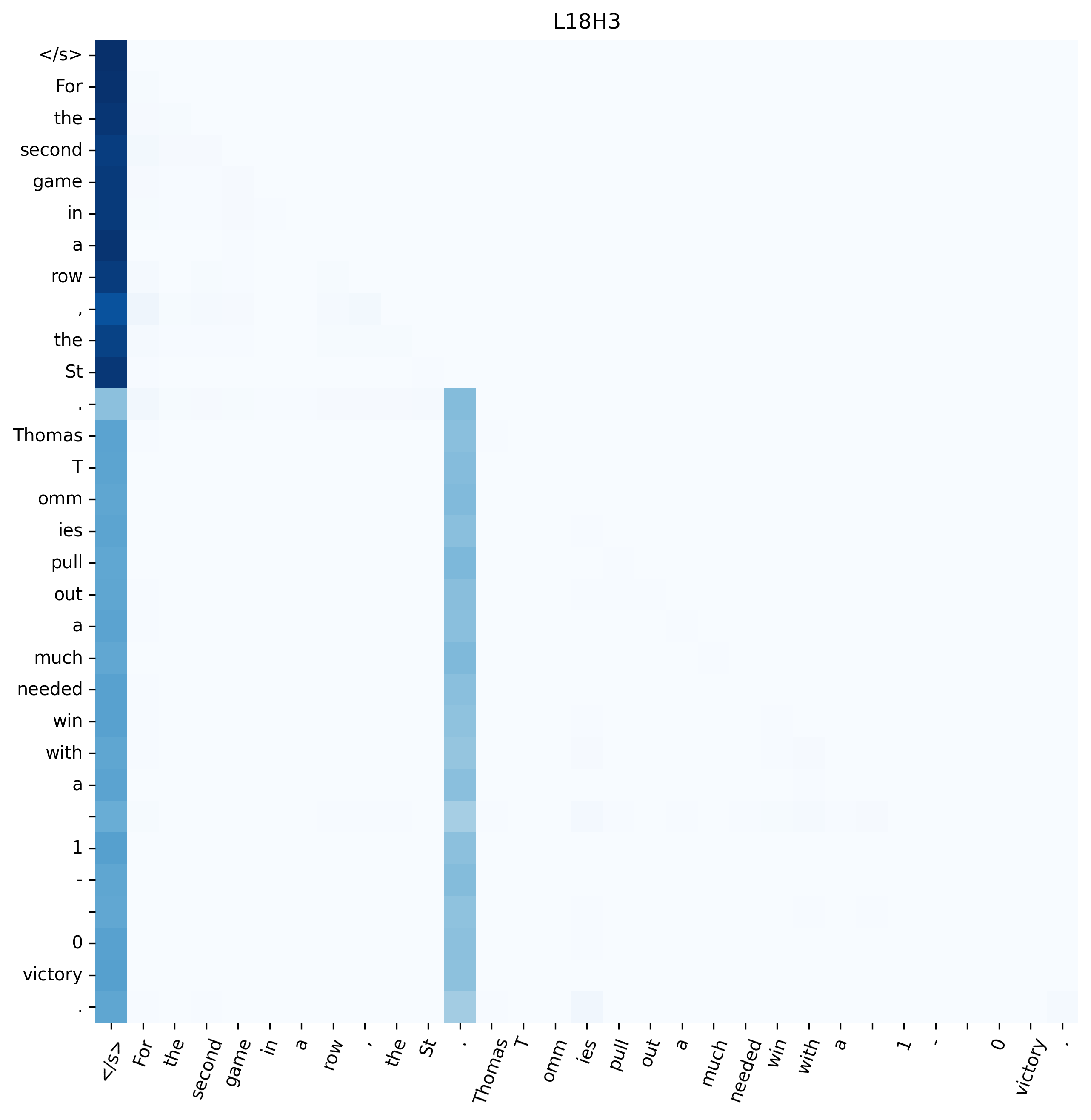}
	\caption{Llama 2 7B attention weights matrix L18H3.}
\label{fig:llama2_7b_special_token_attn_weights_L18H3}
	\end{centering}
\end{figure}

Figure~\ref{fig:llama2_7b_special_token_attn_weights_L18H3} shows an example of an attention map for one of the heads in Llama 2-7b. We see that after the first period, attention is spread between the BOS token and this period.

\end{document}